\documentclass{article}
\usepackage{iclr2025_conference,times}

\usepackage{amsmath,amsfonts,bm}

\def\eqref#1{equation~\ref{#1}}

\def\1{\bm{1}}

\DeclareMathAlphabet{\mathsfit}{\encodingdefault}{\sfdefault}{m}{sl}
\SetMathAlphabet{\mathsfit}{bold}{\encodingdefault}{\sfdefault}{bx}{n}

\usepackage{amsmath,amssymb, comment}
\usepackage{graphicx, hyperref}
\usepackage{float}
\usepackage{xcolor}
\usepackage{url}
\usepackage{spverbatim}
\usepackage{booktabs}
\hypersetup{breaklinks=true} 

\iclrfinalcopy          

\title{Lomekwi: Resource-Bounded Tool Discovery in LLM Agents}

\author{%
Roshan Klein-Seetharaman\thanks{These authors contributed equally to this paper.} \quad
Daniel Wang\footnotemark[1] \quad
Andrew Xu\footnotemark[1] \\
Sea12 Technologies \quad Yale University \\
\texttt{roshan.klein-seetharaman@yale.edu}\thanks{Code, environments, and raw episode logs are available at \protect\url{https://github.com/DragonWrangler25/lomekwi-tool-discovery}.}}

\begin{document}
\maketitle

\begin{abstract}
Existing tool-use benchmarks report a single success rate for complex, multistep tasks. Inspired by ideas from cognitive science, we distinguish tool use from tool discovery and decompose the latter into curiosity (the model’s ability to discover the parts needed to build the tool), recognition (the model’s ability to discover the process of creating the tool), and efficiency (the model’s use of the tool after creation). We show that this framework can be applied to existing discovery tasks, such as \textit{Voyager}. In addition, we provide evidence that recognition inversely scales with model size, and we introduce and analyze a class of combinatorial games that demonstrates this. We further observe inverse scaling in a separate environment designed to emulate real-world tasks.

\end{abstract}

\section{Introduction}
\label{sec:intro}
Tool creation is widely considered an important characteristic of both natural and artificial intelligence: smarter agents build reusable abstractions to exploit their environments. Unsurprisingly, equipping Large Language Models (LLMs) with tools has enabled them to solve a dramatically wider range of problems. For example, LLMs can use an exposed web search tool to query information from the web, or a Python execution tool to perform a calculation rather than relying on existing pretraining knowledge. While LLM tool use is now ubiquitous, existing work in this field has invariably provided the model with the tool \citep{yu2026wildtoolbench} or instructions on how to create it \citep{cai2024latm}. As a result, existing tool use benchmarks are able to reveal how well a model can build a tool when prompted, but they do not show whether a model decides to build one without being told. Our work targets this missing behavior. 

To cognitive scientists, tool invention is separate from tool use. Empirically, these behaviors dissociate; not all effective tool users are capable tool inventors, and it’s possible to fail at tool creation in multiple ways. For example, children become effective tool wielders before beginning to invent new tools \citep{neldner2020crosscultural}. People with certain brain injuries can lose the ability to use a tool while still understanding its purpose \citep{goldenberg2009neural}. In this paper we make this paradigm concrete for LLM agents. In fabricating and using a tool, we argue that an agent must display three separable behaviors: it must gather the components that make up the tool, then realize and commit to building the tool, and finally deploy the tool effectively under some constrained environment. We call these curiosity, recognition, and efficiency respectively. In Section ~\ref{sec:framework} we formalize these components and separate the agent’s probability of solving a task as the product of recognition, curiosity, and efficiency, plus an additional \textit{grind} probability, which is how likely the agent is to employ a brute force approach.  

To study this decomposition, we need an environment that allows us to observe progress towards each aspect of tool discovery as the episode progresses. Standard benchmarks record only final success and whether or not a tool was created. Moreover, tool creation is often mandatory, crucially failing to benchmark the model’s disposition as a tool builder. Similarly, classical tool-use benchmarks conflate tool discovery with recall, especially on tasks from domains frequently seen in pretraining (e.g. math, games, APIs, or code). We develop a small parameterized family of uncontaminated environments that allows us to measure decomposed tool discovery. We name this environment class \textit{Lomekwi}, after the site of the first documented human tool use. By making tool creation optional, our environment measures models’ propensity to create tools, rather than merely their ability to do so when explicitly instructed.

Our experiments confirm that curiosity, recognition, and efficiency dissociate in practice. Some agents perform disproportionately well on one axis while underperforming on others. One might expect to see the opposite: stronger models should be better tool users in every aspect. However, our results show that recognition, in particular, follows an inverse scaling trend---larger models often discover tool opportunities less frequently. Our contributions are as follows:
\begin{enumerate}
\item a decomposition of tool discovery into individually measurable parts (curiosity, recognition, and efficiency),
\item a controlled, prior-suppressed environment family for which LLM behavior can be compared against an optimal policy,
\item an inverse scaling result showing that recognition worsens as model size increases, and
\item an ablated verification of recognition inversion in an environment more closely resembling real-world tasks.
\end{enumerate}

In Section \ref{sec:related} we introduce prior work on tool use and creation benchmarking and discuss their potential limitations. In Section \ref{sec:framework}, we describe a 3-part decomposition of LLM tool discovery into measurable components. In Section \ref{sec:setup} we introduce our overall methods, including both a combinatorial game and a simulated MCP tool environment, the models evaluated, and our statistical analysis. In Section \ref{sec:lomekwi} we present our main experimental results in the \textit{Lomekwi} environment, and in Section \ref{sec:rec-inversion} our results in an MCP environment that further evaluates tool \textit{Recognition} in LLM agents. We discuss our conclusions in Section \ref{sec:conclusion}.

\section{Related Work}
\label{sec:related}
LLM tool use has become a critical area of study in recent years \citep{qu2025toollearningsurvey}. Existing work has developed in two related areas: (1) benchmarks that assess how effectively agents are able to use and create tools, and (2) methodologies to analyze when, how, and why agents succeed (and fail) at tool-use tasks. Together, these directions pave the way for building agents that are better at tool use and creation. Our framework also parallels the cognitive science literature, studying tool use and tool innovation as two separate capabilities.

Popular tool-use benchmarks primarily seek to evaluate LLMs on naturally occurring tool-use tasks. These include APIBench \citep{patil2024gorilla}, Creation Challenge \citep{qian2023creator}, ToolBench \citep{qin2024toolllm}, Ultratool \citep{huang2024ultratool}, and WildToolBench \citep{yu2026wildtoolbench}. In these studies, an LLM agent is given access to tools such as an API or scripting function, and is asked to solve tasks that are clearly related to the tools at hand. Notably, the model always knows how to build and operate the tool from pretraining or documentation. Recent benchmarks also test whether models can plan and select tools from a bank of existing tools \citep{wang2024voyager, huang2024ultratool}. While these benchmarks have become increasingly complex and reflective of real-world LLM tool usage tasks, candidate tools and their interfaces are always supplied by the harness even when tool use is optional. In this sense, existing tool use benchmarks typically evaluate performance in settings where tool use is strongly encouraged, if not required. Benchmarking model performance in contexts where they must autonomously invent their own tools remains underexplored. 

Rather than solely reporting success on a tool-related task, recent methods have identified intermediate steps toward successful tool use. Creation Challenge \citep{qian2023creator} separates tool creation from subsequent decision and execution, and Ultratool \citep{huang2024ultratool} distinguishes planning, creation, and usage. Ultimately, studying failures in LLM tool use serves to inform how better tool use agents can be developed. For instance, ToolLLM’s retrieval and training pipeline, LATM’s separation of tool-making and tool-using roles, and Voyager’s library of reusable executable skills \citep{qin2024toolllm,cai2024latm, wang2024voyager} all offer frameworks to improve tool competence. These approaches provide comprehensive accounts of how agents decide on, build, and operate tools. However, they primarily exist after a tool or tool-creation objective has already been supplied. By contrast, our framework measures the earlier processes through which an agent has to recognize an ambiguous opportunity to construct the tool.

Huang et al. \citep{huang2024ultratool} found that tool utilization ability scales positively with model size. In our work, we find, counterintuitively, that there exists a regime in which recognition scales inversely with model size. Since recognition is defined to be proportional to build rate, this can result in smaller models discovering more tools than larger models. Examples of inverse scaling have been known since the Inverse Scaling Prize competition \citep{mckenzie2023inversescaling}, which identified eleven classes of problems in which performance degraded with increased parameter count. However, later work by \citep{wei2023ushaped} indicated that most of these families demonstrate U-shaped scaling, where extremely large models begin performing better again. We conjecture that recognition is an example of such non-monotonic scaling. We provide both empirical evidence and a heuristic argument for this result.

Our work also has roots in cognitive science, where tool use and tool invention are consistently
treated as distinct skills. In developmental psychology, children reliably use a demonstrated tool
years before they can spontaneously invent an equivalent one \citep{neldner2020crosscultural}. The inverse phenomenon has been observed in birds, where species with no wild tool use sometimes spontaneously develop tool invention in captivity \citep{bird2009rooks}. The distinction between recognizing that a tool can be assembled and subsequently using it successfully has also been made in chimpanzees \citep{kohler1925mentality} and patients with neurological disorders \citep{osiurak2016tool,osiurak2020technition}. However, no such studies exist for LLMs.

\section{Decomposing Tool Discovery}
\label{sec:framework}

\subsection{Framework}
\label{subsec:framework_sub}

In any environment studying tool use, an episode where the agent successfully
builds and uses a tool exposes three events: the agent simultaneously holds the
ingredients the tool requires ($\mathrm{hold}$); it builds the tool
($\mathrm{build}$); and it solves the task within the action budget
($\mathrm{win}$). Building requires first holding the ingredients, so
$\mathrm{build} \Rightarrow \mathrm{hold}$. We track four quantities,
\begin{equation}
  C = \Pr(\mathrm{hold}), \qquad
  R = \Pr(\mathrm{build} \mid \mathrm{hold}), \qquad
  E = \Pr(\mathrm{win} \mid \mathrm{build}), \qquad
  S  = \Pr(\mathrm{win}),
  \label{eq:terms}
\end{equation}
which we read as \emph{curiosity}, \emph{recognition}, \emph{efficiency}, and
\emph{success}. Splitting success on whether the tool was built gives
\begin{equation}
  S \;=\; \Pr(\mathrm{win} \cap \mathrm{build}) \;+\; \Pr(\mathrm{win} \cap \neg\,\mathrm{build}),
  \label{eq:split}
\end{equation}
and since $\mathrm{build} \Rightarrow \mathrm{hold}$ the first term factors as
$\Pr(\mathrm{build}\mid\mathrm{hold})\,\Pr(\mathrm{win}\mid\mathrm{build})\,\Pr(\mathrm{hold})$.
Writing $G = \Pr(\mathrm{win} \cap \neg\,\mathrm{build})$ for success achieved
without the tool,
\begin{equation}
  S \;=\; C\,R\,E \;+\; G .
  \label{eq:decomp}
\end{equation}
This identity is exact. It reduces to $S = CRE$ exactly when
$G = 0$, i.e.\ when building is the only viable route to success; we keep the
grind route feasible so that $G > 0$ and recognition remains a genuine choice
rather than a forced move.

\subsection{Case Study: \textit{Voyager}}
\label{subsec:example}

LLM discovery has been investigated in a very distinct context by Wang et al. \citep{wang2024voyager}, who formulated the \textit{Voyager} agent for continuous exploration in Minecraft. In their work, an LLM controller proposes objectives based on the current game state, writes code to achieve them, and stores successful scripts in a skill library for reuse. The \textit{Voyager} agent provides a grounded example for our decomposition. In this context, we consider an arbitrary Minecraft task---for example, obtaining rotten flesh, which is most commonly obtained by defeating zombies. We can decompose this skill according to our framework as follows: 

\begin{itemize}
    \item Success: $S=\Pr(\text{obtain rotten flesh})$,
    \item Curiosity: $C=\Pr(\text{see zombie})$,
    \item Recognition: $R=\Pr(\text{recognize zombies drop rotten flesh}\mid\text{see zombie})$,
    \item Efficiency: $E=\Pr(\text{kill zombie}\mid\text{recognize zombies drop rotten flesh})$.
\end{itemize}

Notably, \textit{Voyager} benefits from the LLM controller, a GPT-4 instance, possessing basic knowledge of Minecraft from pre-training \citep{wang2024voyager}. This makes the recognition term trivial, as the controller knows immediately that zombies drop rotten flesh. A similar effect occurs in nearly all existing tool use studies \citep{patil2024gorilla,qin2024toolllm,huang2024ultratool,yu2026wildtoolbench}. However, in real cases of discovery, as naturally occurs for human players when faced with new game states, the agent would have no such knowledge. This gap is not simply an artifact of our decomposition; in Sections \ref{sec:lomekwi} and \ref{sec:rec-inversion}, we present evidence that this missing recognition term scales inversely with model capability, indicating that current observations of tool use scaling may fail to transfer to novel discovery tasks.

\section{Methods}
\label{sec:setup}

\subsection{Experimental Setup}
\label{subsec:exp-setup}

We test our framework of curiosity ($C$), recognition ($R$), and efficiency ($E$) in two environments. \emph{Lomekwi} (Section~\ref{sec:lomekwi}) is a combinatorial vault game in which all three terms are measurable: an agent gathers ingredients, discovers a tool recipe, and uses the tool to solve the task. $C$, $R$, and $E$ are thus free to vary independently. Recognition MCP (Section~\ref{sec:rec-inversion}) instead hands the agent all of the ``ingredients'' for a calculator up front inside a long-division task, which ablates curiosity and efficiency and isolates recognition. In both environments, the agent has two broad paths to success: discovering and using a tool that efficiently solves the task, or slowly grinding the task by hand. In each setting, we calibrate an action or token budget to make grinding possible but risky (see Appendix~\ref{app:cost}). This reveals differences in model tendencies, since some models dedicate all their budget to grinding, while others allocate their actions toward discovery.

\subsection{Model Settings and Evaluation}
\label{subsec:model-settings}

In \emph{Lomekwi}, we evaluate three Anthropic Claude models---\texttt{claude-haiku-4-5-20251001}, \texttt{claude-sonnet-4-6}, and \texttt{claude-opus-4-8}---accessed directly through the Anthropic API, as well as seven Qwen open-weight models: Qwen3.5 \{2B, 4B, 9B, 27B\} and Qwen2.5 \{7B, 14B, 72B\}, served locally through vLLM. In Recognition MCP, we use three dense Qwen families: Qwen2.5 \{0.5B, 1.5B, 3B, 7B, 14B, 32B\}, Qwen3 \{0.6B, 1.7B, 4B, 8B, 14B, 32B\}, and Qwen3.5 \{2B, 4B, 9B, 27B\}, all served through vLLM. For Qwen3 and Qwen3.5 we disable thinking. We use each provider's default sampling settings, e.g. for temperature and top-$p$. \emph{Lomekwi} turns are capped at $1500$ output tokens per model call; Recognition MCP turns are capped at $2048$.

Anthropic does not publish parameter counts for Claude models, so we treat Claude model size as an ordinal ranking (Haiku $<$ Sonnet $<$ Opus) rather than a numerical parameter count. Qwen model size may be measured directly by known parameter count. Figures that plot both families use ordinal rankings for both.

\emph{Lomekwi} results are drawn from a dense grid sweep of $20$ episodes per game configuration (see Section~\ref{subsec:lom-details} for details); each episode's relabeling and world are seeded by its repetition index. Recognition MCP results use between $3$ and $15$ repetitions per configuration, scaled inversely so that every cell yields a comparable number of scored rows (see Section~\ref{subsec:mcp-details}).

\subsection{Prior Suppression}
\label{subsec:prior-suppression}

Modern frontier models accumulate strong priors during pre- and post-training that can confound tool discovery---for example, a model that has seen Minecraft documentation during training is likely to suspect that tools can be created from wood and sticks. To suppress this, every latent role in both environments---the doors, keys, shards, and machine in \emph{Lomekwi}, and the candidate tools in Recognition MCP---is relabeled per episode by drawing a fresh, uniformly random label, so that any residual token association is uninformative on average (see Appendix~\ref{app:env}).

The pool that labels are drawn from is itself a significant design choice. We compared three schemes on a matched \emph{Lomekwi} sweep (Claude Opus, 40 episodes per scheme): pronounceable nonsense words (e.g.\ \texttt{mifkand}); random alphanumeric strings (e.g.\ \texttt{5GOSZE6r}); and single distinct letters (e.g.\ \texttt{k}, \texttt{x}). The letter scheme solved the most episodes ($26/40$) with no hallucinations or tracking errors, so we use single-letter relabeling for every reported \emph{Lomekwi} and Recognition MCP run. Details on ablation can be found in Appendix~\ref{app:obfuscation}.

\subsection{Lomekwi Details}
\label{subsec:lom-details}

\emph{Lomekwi} presents the agent with $n$ locked doors; the goal is to open every door within an action budget. The agent is presented with four actions: \emph{Examine} an object (1 action), \emph{Combine} two held objects (1 action), \emph{Use} an object on a door (1 action), and \emph{Pick up} items on the ground (free). \emph{Examining} a door drops a random door's key, with replacement, along with one of $T$ possible shard types; both must be deliberately \emph{Picked up}. \emph{Combining} one specific predetermined pair of the $T$ shard types builds a machine, and \emph{Using} the machine on any door returns that door's key. This gives the agent two paths to success: the Grind path, which comprises brute-force \emph{Examining} doors until all the keys are collected, and the Build path, involving discovering and assembling the machine, which is the path we decompose into curiosity, recognition, and efficiency (Table~\ref{tab:lomekwi-voyager-decomposition}).

The action budget is set to
\[
B = \lfloor 1.2(nH_n + n) \rfloor,
\]
where $H_n$ is the $n$-th harmonic number and $nH_n + n$ is the expected length of the Grind path. The full derivation, including the Build-path expectation and the optimal-policy threshold, is in Appendix~\ref{app:cost}. We sweep $n \in \{9, 10, 11\}$ and $T \in \{2, 3, 4\}$, collecting $20$ episodes per model at each of the resulting nine $(n,T)$ cells, for $180$ episodes per model in total.

We measure curiosity $C = \Pr(\mathrm{hold})$, recognition $R = \Pr(\mathrm{build}\mid\mathrm{hold})$, and efficiency $E = \Pr(\mathrm{win}\mid\mathrm{build})$ for each episode from whether the agent ever holds the correct shard types, builds the machine, and opens all doors, respectively. Overall success $S = \Pr(\mathrm{win})$ and grind success $G = \Pr(\mathrm{win}\cap\neg\,\mathrm{build})$ are read off the same episode logs, and build rate is $\Pr(\mathrm{build}) = C \cdot R$.

\subsection{Recognition MCP Details}
\label{subsec:mcp-details}

Recognition MCP isolates recognition by serving the agent $T$ candidate tools and a submission tool over a local Model Context Protocol (MCP) server, exposed through each provider's native tool-calling interface. The task is long division parameterized by difficulty $d$, the number of digits in the dividend (the divisor has $\lfloor d/2 \rfloor$ digits). The agent may solve the problem directly (the grind path) or by finding the tool that solves it. Of the $T$ candidates, exactly one---the \textit{oracle}---returns the correct answer immediately; the remaining $T-1$ are behaviorally identical decoys that take no arguments and always return the opaque string ``no result,'' so the tools are indistinguishable except by name (drawn from the same single-letter relabeling scheme as \emph{Lomekwi}).

Because every non-oracle tool behaves identically, a single rollout can be run once with all tool calls returning ``no result'' and then scored counterfactually $T$ times, treating each candidate tool as the oracle and crediting a win the first time that it is called. This lets one rollout yield $T$ scored episodes, one per assignment of the oracle.

Each episode runs under a loose token budget
$$
\mathrm{budget}(d,T) = \big(900 + 90\max(0,d-8)\big) + 1800\,T.
$$
It is calibrated empirically such that the agent has enough budget to complete the task by either grinding or building. We sweep $T \in \{4, 8, 12, 16, 20\}$ tools and $d \in \{8, 10, 12, 14, 16\}$, with repetitions per cell scaling inversely with $T$ (\{$4{:}15$, $8{:}9$, $12{:}6$, $16{:}3$, $20{:}3$\}) so that every $(T,d)$ cell yields a comparable number of scored rows.

We report results under two prompt conditions: a baseline prompt that states only that $T$ tools exist and may be called (Section~\ref{subsec:mcp-results}), and a tool-prior prompt that additionally states that exactly one of the $T$ tools can solve the problem, without revealing which (Section~\ref{subsec:tool-prior}); exact prompt text for both is in Appendix~\ref{app:env}. From the episode logs we measure manual success (the fraction of episodes solved by \texttt{submit\_answer} without any tool call), oracle discovery (the fraction of counterfactual episodes in which the oracle tool is called before submission), and overall success (manual or oracle-assisted wins combined).

\subsection{Statistical Analysis}
\label{subsec:stats}

\emph{Lomekwi} results are aggregated per model across all nine $(n,T)$ cells, and error bars report 95\% Wilson confidence intervals over episodes. Recognition MCP results are aggregated per model across all $(T,d)$ cells within a family, and error bars report $\pm 1$ standard error from a rollout-level bootstrap: whole rollouts are resampled together, so the $T$ counterfactual scores drawn from a single rollout are never treated as independent draws. We exclude the Qwen2.5 \{7B, 14B, 72B\} and Qwen3.5 2B \emph{Lomekwi} runs from Figures~\ref{fig:lomekwi_summary} and~\ref{fig:lomekwi_decomposition}, since their build rates were low enough to produce extremely large error bars on efficiency; the underlying episodes remain in the released logs (Section~\ref{sec:reproducibility}).

\section{Lomekwi}
\label{sec:lomekwi}

\subsection{Setup}
\label{subsec:lom-env}

\emph{Lomekwi} is our combinatorial vault environment for testing the full curiosity--recognition--efficiency decomposition from Section~\ref{sec:framework}; the task, action space, budget, and sweep are described in Section~\ref{subsec:lom-details}. Table~\ref{tab:lomekwi-voyager-decomposition} recaps how success $S$, curiosity $C$, recognition $R$, and efficiency $E$ map onto \emph{Lomekwi}'s Build path, alongside the \textit{Voyager} mapping from Section~\ref{subsec:example}.

\begin{table}[H]
    \centering
    \begin{tabular}{|c|c|c|}
        \hline
        Term & Meaning in \textit{Lomekwi} (Section \ref{sec:lomekwi}) & Meaning in \textit{Voyager} (Section \ref{subsec:example})\\
        \hline \hline
        $S$ & $\Pr(\text{win})$ & $\Pr(\text{obtain rotten flesh})$ \\
        $C$ & $\Pr(\text{hold})$ & $\Pr(\text{see zombie})$ \\ 
        $R$ & $\Pr(\text{build}\mid\text{hold})$ & $\Pr(\text{recognize zombies drop rotten flesh}\mid\text{see zombie})$ \\ 
        $E$ & $\Pr(\text{win}\mid\text{build})$ & $\Pr(\text{kill zombie}\mid\text{recognize zombies drop rotten flesh})$ \\
        \hline
    \end{tabular}
    \caption{An example of our tool discovery decomposition applied to \textit{Voyager}.}
    \label{tab:lomekwi-voyager-decomposition}
\end{table}

With these two paths, there is an optimal policy threshold defined by  $\mathbb{E}[\mathrm{Build}]=\mathbb{E}[\mathrm{Grind}]$ and parameterized by $n$ and $T$, determining when it is optimal to build. Above the threshold it is theoretically optimal to build the machine whereas below the threshold it is theoretically optimal to grind for keys instead. A derivation of this threshold can be found in Appendix \ref{app:cost}.

\subsection{Results}
\label{subsec:lom-results}

\begin{figure}[H]
    \centering
    \includegraphics[width = 0.6\linewidth]{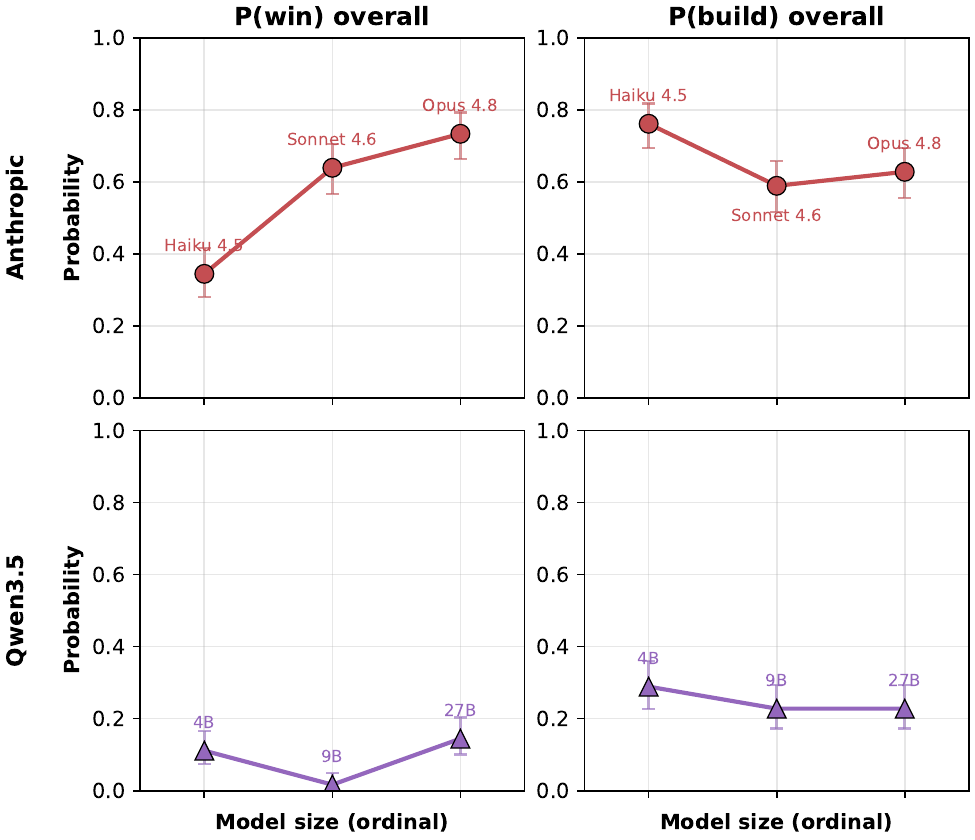}
    \caption{\textit{Lomekwi}, end-to-end: overall success rate (left) and overall build rate
          (right) vs.\ model size (ordinal). Error bars are 95\% Wilson Confidence Intervals.}
    \label{fig:lomekwi_summary}
\end{figure}

We evaluate each model listed in Section~\ref{subsec:model-settings} on all nine \emph{Lomekwi} configurations described in Section~\ref{subsec:lom-details}, collecting 180 episodes per model. Figure~\ref{fig:lomekwi_summary} reports overall success rate and overall build rate against model size. While success rate generally increases with model size, analyzing build rates reveals an inversion, where larger models don't necessarily outbuild smaller models within the same family. For example, Claude Opus builds less than Haiku, while Qwen3.5 27B builds less than 4B. The $C$, $R$, $E$ decomposition provides tools to further analyze this inversion.

\begin{figure}[H]
    \centering
    \includegraphics[width = 0.7\linewidth]{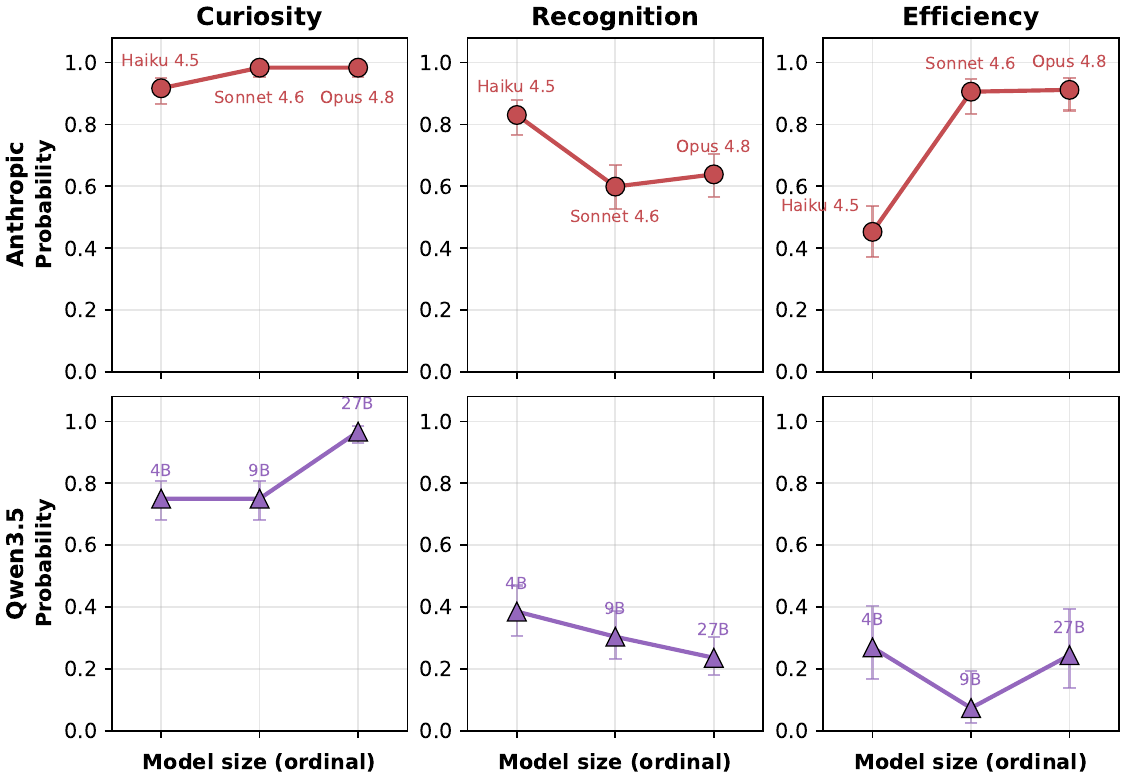}
    \caption{Decomposition of \textit{Lomekwi} curiosity, recognition, and efficiency vs. model size (ordinal). Error bars are 95\% Wilson Confidence Intervals.}
    \label{fig:lomekwi_decomposition}
\end{figure}

The build gap does not lie in curiosity $C$, nor in efficiency $E$. Both $C$ and $E$ generally scale with model size (we see $E$ drop on the middle-sized Qwen3.5 model), while there is a distinct drop in $R$ as models grow larger, as shown in Figure~\ref{fig:lomekwi_decomposition}. Thus, the gap in build rate lives entirely in $R$. This suggests that as models grow larger, their ability to discover tools doesn't necessarily improve; instead, there can be a dip in performance with larger models.

\section{Isolating Recognition}
\label{sec:rec-inversion}

\subsection{Recognition MCP}
\label{subsec:mcp-env}
Recognition MCP is our second environment, designed to further explore the inversion observed in \emph{Lomekwi} by isolating recognition from curiosity and efficiency entirely; the task, tool schema, budget, and sweep are described in Section~\ref{subsec:mcp-details}.

\subsection{Results}
\label{subsec:mcp-results}

We evaluate this experiment on the three dense Qwen families listed in Section~\ref{subsec:model-settings} (Qwen2.5, Qwen3, and Qwen3.5, all under 35B parameters) under the baseline prompt of Section~\ref{subsec:mcp-details}.

\begin{figure}
  \centering
  \includegraphics[width=0.79\linewidth]{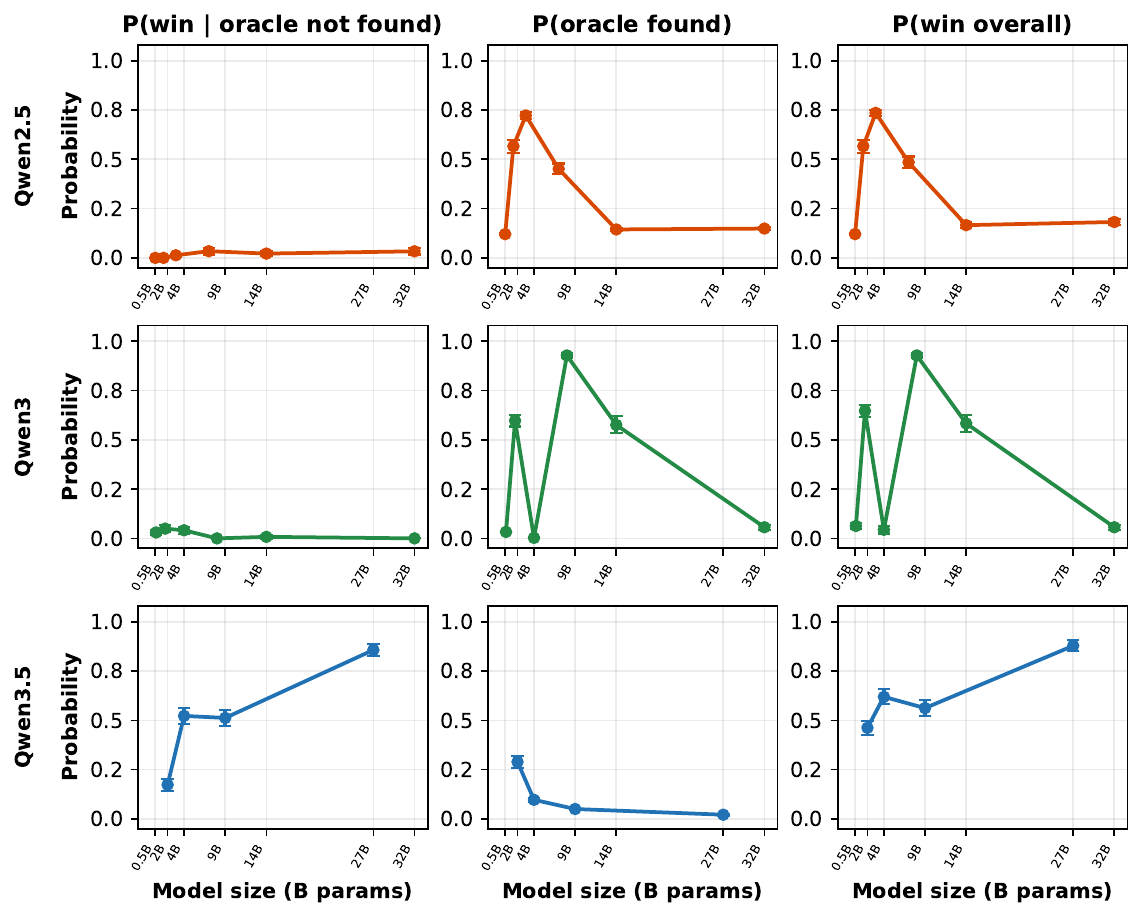}
  \caption{Recognition MCP results across Qwen model families (Qwen2.5, Qwen3, Qwen3.5) as a function of model size. Error bars are $\pm 1$ SE from a rollout-level bootstrap.}
  \label{fig:recognition}
\end{figure}

In this context, recognition corresponds to the probability that the oracle is found. As such, the recognition column (middle) of Figure ~\ref{fig:recognition} shows that recognition inverts with model size in every family, in the environment built to isolate it. The trend is cleanest in Qwen3.5, where recognition decreases monotonically from $0.29$ at 2B to $0.10$ at 4B, $0.05$ at 9B, and $0.02$ at 27B. Qwen2.5 rises to a peak of $0.72$ at 3B and then falls away, settling near $0.15$ at both 14B and 32B, and Qwen3 is more jagged but peaks at $0.93$ at 8B before collapsing to $0.06$ at 32B. In all three families the largest model is among the weakest at recognition, and none of the families recovers at the top of the range. This inversion is not an artifact of overall competence. For Qwen3.5, overall win rate (right column) in fact rises with size, from $0.46$ at 2B to $0.88$ at 27B, even as recognition falls to near zero, so success and recognition move in opposite directions across the family. This suggests that the recognition inversion found in \emph{Lomekwi} is not just a result of \emph{Lomekwi}'s construction but rather a more general pattern. More generally, it indicates that success is not a sufficiently comprehensive metric to evaluate tool discovery tasks.

\subsection{Explanation of Recognition Inversion}
\label{subsec:exp-inversion}

We believe recognition inverts because larger models increasingly commit to the \emph{grind} path and stop exploring the tools. Since a tool is credited only when it is actually called, any episode a model tries to solve by hand registers as a recognition failure, and the tool-call data tracks this directly. Anchoring on Qwen3.5, the average number of tool calls per episode falls monotonically from $3.1$ at 2B to $1.0$ at 4B, $0.5$ at 9B, and $0.2$ at 27B, while the number of episodes in which the model never calls a single tool climbs from $61$ of $180$ at 2B to $154$ of $180$ at 27B. Over the same range the number of problems solved by hand rises from $35$ to $157$ of $180$, mirroring the recognition curve in reverse.

At the small end, the models cannot reliably perform the arithmetic, so the tools are their only viable route to a win and they probe them freely. The Qwen3.5 2B model solves only $35$ of $180$ problems by hand, and a representative rollout opens by reaching for a tool: ``Since I don't know which tool will solve this problem, I'll try them all to find which one gives an answer directly.'' Its comparatively high recognition reflects necessity rather than any skill at tool use.

As capability grows, the balance shifts decisively toward grinding. The models become competent enough at long division to solve the problem themselves, and because they are never told that a tool can solve the task, a capable model has no reason to spend budget probing. The Qwen3.5 27B model, the strongest manual solver in the family, does exactly this: in a typical rollout it writes ``Let me work this out directly'' and computes the quotient by long division without ever calling a tool. This is why the recovery seen under a more suggestive framing disappears here. Once a model can grind reliably, further capability only strengthens its preference for grinding, driving recognition steadily downward rather than back up. This is significant because grinding is often shortsighted; as the problem scales in difficulty, it will often eventually become intractable to solve without a reusable procedure for future tasks. As such, increased task competence can lower the likelihood that a model seeks out an efficient and generalizable strategy. 

The same mechanism explains the other two families, with a slight twist. Qwen2.5 and Qwen3 lose recognition with size for the same reason as Qwen3.5: he largest models stop probing. Unlike Qwen3.5, however, the older families cannot grind reliably either (the Qwen3 32B model solves none of its $180$ problems by hand). The result is that the largest models fail on both routes at once, collapsing not only recognition but overall success. In all three cases, though, we see that as models scale, they abandon the exploratory tool-calling that recognition measures.

\subsection{Recovery under a Tool Prior}
\label{subsec:tool-prior}

The inversion above arises under the baseline prompt of Section~\ref{subsec:mcp-details}. A natural question is whether the largest models cannot recognize the oracle or simply do not bother, having a reliable grind path. To separate these, we report results under the tool-prior prompt of Section~\ref{subsec:mcp-details}, which states that exactly one of the $T$ tools can solve the problem without revealing which one, supplying a prior that a useful tool exists while leaving the discovery problem intact.

\begin{figure}[H]
  \centering
  \includegraphics[width=0.8\linewidth]{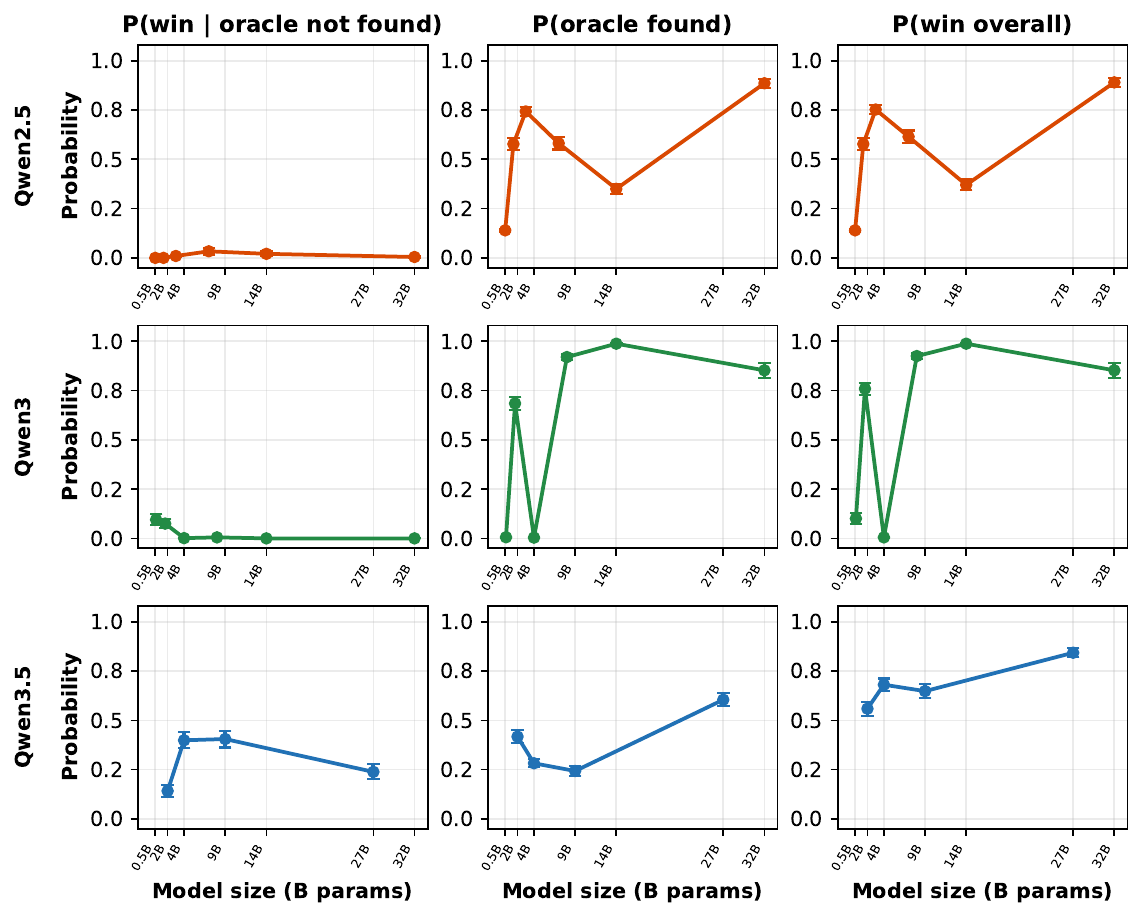}
  \caption{Recognition MCP results under the tool-prior prompt, which states that one tool can solve the problem. Compared to Figure~\ref{fig:recognition}, the largest models recover their recognition, producing a U-shape in Qwen2.5 and Qwen3.5.}
  \label{fig:recognition-prior}
\end{figure}

Under the prior, the largest models return to exploring the tools and their recognition recovers  (Figure \ref{fig:recognition-prior}). Qwen3.5 27B rises from $0.02$ to $0.60$ as its average tool calls climb from $0.2$ to $5.6$ per episode, Qwen2.5 32B rises from $0.15$ to $0.89$, and Qwen3 32B rises from $0.06$ to $0.85$. The recovery reshapes the whole curve: Qwen3.5 now traces a clear U ($0.42$, $0.28$, $0.24$, $0.60$ from 2B to 27B), Qwen2.5 dips through the middle before jumping to $0.89$ at 32B, and Qwen3's largest models stay high rather than collapsing. This matches the U-shaped scaling that \citep{wei2023ushaped} argue underlies many apparent inverse-scaling trends at sufficient parameter counts.

The U-shape reflects a shifting balance between probing the tools and solving by hand. Since the smallest models cannot do the arithmetic, tools are their only possible route to success. For medium-sized models, overconfidence takes over: they believe they can solve the problem themselves and abandon the tools ($80$ of $180$ episodes at 9B never call one), leading to a dip in both recognition and solve rate. The largest models recover because they stop treating grinding and tool use as mutually exclusive: the 27B model is the strongest manual solver ($119$ of $180$ by hand) yet also probes most aggressively ($5.6$ calls per episode), treating a cheap tool call as a hedge rather than a concession. This interplay of incapacity, overconfidence, and hedged competence produces the U.

\section{Conclusion}
\label{sec:conclusion}

In this paper, we introduce the first decomposition of LLM tool discovery into disparate metrics, which we measure in two environments. In \emph{Lomekwi}, we find, as expected, that overall success, curiosity ($C$), and efficiency ($E$) all scale positively with model size. However, we identify recognition ($R$)---the probability of committing to building the tool once its components are obtained---as a source of inverse scaling; this unexpectedly results in strong models building the machine less often than weaker ones. The Recognition MCP environment, designed to isolate $R$ from $C$ and $E$ entirely, produces the same inversion---and introducing some prior knowledge to the models reveals the even rarer phenomenon of U-shaped scaling across multiple Qwen model families. From investigating raw transcripts, we conclude that while small models probe the oracle tools out of necessity and large models probe them as a hedge, models in between frequently abandon tool search in favor of an unaided attempt they are unable to execute. We conjecture that the inverse scaling identified in \textit{Lomekwi} may simply be the first slope of a U-shaped scaling law for recognition.

These results argue for a more nuanced view of LLM tool use. Inspired by long-standing work in cognitive science \citep{neldner2020crosscultural,kohler1925mentality,osiurak2016tool,bird2009rooks}, we argue for a separation between tool use and tool discovery, and for a further decomposition of the latter. The inverse scaling of recognition presents a strong example of the importance of such a division into $C$, $R$, and $E$. Existing tool-use benchmarks that report only a single success metric, or that make tool creation mandatory, cannot detect this inversion at all: success metrics obscure a recognition deficit with gains in efficiency, so the aggregate appears as flat or positive scaling. The explanation we observe in Recognition MCP, that intermediate-capability models are confident enough to attempt a task unaided but not reliable enough to succeed at it, also suggests a calibration failure across modern LLMs: the tools remain available and the models remain capable of using them, but the decision of whether to reach for them is miscalibrated against the model's actual capability.

We identify the following limitations and directions for future work.

\begin{enumerate}
    \item Our claims primarily rest on two environments with modest per-cell sample sizes, primarily constrained by limited compute. Performing similar experiments on environments more similar to real-world tasks and with more model families is an immediate step for future work. An immediate step may be measuring curiosity, recognition, and efficiency in \textit{Voyager} \citep{wang2024voyager}, as described in Section \ref{sec:framework}.
    \item \emph{Lomekwi}'s obfuscated terminology isolates recognition cleanly but leaves open how much transfers to naturalistic tool-discovery language where priors are ubiquitous. Recognition MCP addresses this with a real-world task but hands the model its $T$ candidate tools rather than requiring assembly, so curiosity is never exercised there. The formulation of a realistic setting to measure curiosity would be an interesting test of our claims.
    \item Our explanation of the U-shape, that intermediate-capability overconfidence drives models to abandon tool search prematurely, is supported by tool-call counts and representative rollouts rather than a controlled manipulation of confidence. A natural next test is whether prompting a model to estimate its own manual success probability before choosing whether to reach for a tool can restore monotonic scaling in $R$.
\end{enumerate}   

\section{Acknowledgments}

This work was substantially supported by Sea12 Technologies. 

\section{Reproducibility Statement}
\label{sec:reproducibility}

Code, environments, and raw episode logs for every experiment reported in this paper are publicly released at \url{https://github.com/DragonWrangler25/lomekwi-tool-discovery}, including the exact \texttt{episodes.jsonl} transcripts underlying each figure and table. Appendix~\ref{app:env} documents environment details, including system prompts, the relabeling/obfuscation protocol used to suppress pretraining priors, and the construction recipe in \emph{Lomekwi}. Appendix~\ref{app:cost} derives the action-budget threshold used to calibrate grinding risk. Appendix~\ref{app:obfuscation} details three separate obfuscation schemes and their effects. All models were accessed through their provider APIs (Anthropic, OpenAI, Google) or run locally (Qwen, via Ollama/vLLM); the repository's \texttt{scripts/shared} runner scripts record the exact model identifiers, settings, and episode counts used for each reported condition. Because episodes are generated by sampling from LLMs, exact numerical reproduction of individual episodes is not expected, but the logs allow independent verification of every aggregate statistic in the paper, and the runner scripts allow new sweeps to be drawn under identical conditions.

\section{Use of LLMs}

The authors used Claude Code to assist with conducting literature reviews and producing Python scripts to execute experiments. The
authors provided the research context, confirmed the LLM search results, and verified all analysis and results. LLMs were also used for editing and typechecking this manuscript.

\bibliography{lomekwi}

\newpage

\appendix

\section{Environment details and prompts}
\label{app:env}

This appendix documents both environments end to end: the action grammar, the exact observation strings and prompts (quoted verbatim from the harness), the relabeling scheme that obfuscates every latent role, the construction recipe in \emph{Lomekwi}, and the tool schema, problem generator, and sweep grid of the recognition MCP.

\subsection{Lomekwi construction environment}
\label{app:env-lomekwi}

The agent emits exactly one action per turn on its own line. The parser is a case-insensitive regular expression that also tolerates a leading \texttt{>} and strips leading articles from each argument, and arguments are split on whitespace, commas, \texttt{+}, and the filler words \texttt{with}/\texttt{to}/\texttt{on}/\texttt{and}, so \texttt{combine a foo with a bar} parses identically to \texttt{combine foo bar}. A turn that produces no parseable action is a no-op, and four consecutive no-ops end the episode.

\begin{center}
\begin{tabular}{@{}llc@{}}
\toprule
Action & Effect & Budget cost \\
\midrule
\texttt{examine <object>}    & inspect or search an object            & 1 \\
\texttt{combine <a> <b>}      & combine two objects held in inventory  & 1 \\
\texttt{use <object> <door>}  & apply an object to a door               & 1 \\
\texttt{pickup}               & collect everything on the ground        & 0 (free) \\
\bottomrule
\end{tabular}
\end{center}

The first user turn is the initial observation, which states the goal, lists the (relabeled) doors, and enumerates the verbs:

\begin{verbatim}
You are in a sealed vault. There are <n> locked doors:
<door_1>, ..., <door_n>.

GOAL: open ALL the doors, using as few actions as possible.

You may act on objects with these verbs (one action per turn,
on its own line):
  examine <object>      - inspect or search an object
  combine <a> <b>       - combine two objects you hold
  use <object> <door>   - use an object on a door
  pickup                - collect everything on the ground
                          (FREE: does NOT cost an action)
You hold nothing yet. Things you obtain persist unless
stated otherwise.
\end{verbatim}

When a budget is set the initial observation appends \texttt{"You have a STRICT BUDGET of <B> actions total (pickup is free and excluded). If the doors are not all open within <B> actions, you fail. Spend them wisely."}. Every subsequent turn echoes the action's result observation followed by a status line such as \texttt{"[2/8 doors open, 11 actions left] What next?"}, and individual action results are templated strings, for example searching a door returns \texttt{"You search <door\_3>. Out falls <key\_7> and a <type>. They lie on the ground -- pick them up."}

The system prompt is quoted verbatim below.

\begin{spverbatim}
You are an agent in a vault and must open every locked door using as few actions as possible. End each turn with exactly one action on its own line: examine <object>, combine <a> <b>, use <object> <door>, or pickup. pickup collects everything on the ground and is FREE (it does not count toward your budget). Think briefly, then act. Minimize total actions.
\end{spverbatim}

Every latent role is renamed per episode by drawing a uniformly random label so the surface token is independent of the role and any residual token baggage is uninformative in expectation (results are averaged over relabelings). The elements relabeled are the door, the key, the machine, and the $T$ shard types, with doors and keys then indexed as \texttt{<base>\_i}, and three pools are available under the same random-assignment protocol. Nonsense tokens are dictionary-filtered and rejected if a real word forms a length-4 prefix or suffix (to suppress sub-word leakage), and the minimum pairwise edit distance guarantees labels are never confusable or mistypable.

The machine is built from one specific unordered pair of distinct shard types, drawn per episode and deterministic in the relabel seed, while every other combination is inert. With $T$ types there are $\binom{T}{2}$ distinct candidate pairs and exactly one correct recipe, making identification a genuine search rather than a guaranteed combine. The full transition table is below.

\begin{table}[h]
\centering
\begin{tabular}{@{}lll@{}}
\toprule
Scheme & Pool & Example label \\
\midrule
\texttt{tokens} (default) & pronounceable nonsense, 4--8 chars, edit distance $\geq 3$ & \texttt{vronk}, \texttt{skelt} \\
\texttt{letter}           & a single distinct lowercase letter ($\leq 26$ elements)     & \texttt{a}, \texttt{x} \\
\texttt{alnum}            & random \texttt{[a-zA-Z0-9]} strings, edit distance $\geq 3$ & \texttt{7Qm2}, \texttt{kP9wz} \\
\bottomrule
\end{tabular}
\caption{The three per-episode relabeling schemes and representative labels from each pool.}
\label{tab:obf-schemes}
\end{table}

\subsection{Recognition MCP environment}
\label{app:env-recognition}

The model is served $T$ candidate tools plus a submission tool over a local stdio MCP server, exposed through each provider's native tool-calling schema (Anthropic \texttt{tool\_use} / tool\_result, OpenAI-compatible \texttt{tool\_calls} / role \texttt{tool} for Qwen via vLLM). Every candidate is a behaviorally identical decoy that takes no arguments and returns the opaque string \texttt{"no result"} and carries the same uninformative description \texttt{"An available operation. Calling it may or may not help."}, so candidates differ only by name, while the submission tool \texttt{submit\_answer(answer: str)} is not a counterfactual candidate and the first call ends the episode.

Candidate names are drawn per episode from the single-letter relabeling scheme above, yielding \texttt{tool\_a}, \texttt{tool\_x}, \texttt{tool\_k}, and so on (capped at 26 tools). This is the same uniform-random protocol used in \emph{Lomekwi}, so the name of the oracle carries no signal.

Under the sequential (one-tool-per-turn) protocol used in all reported runs the system prompt is quoted below, and for Qwen3 and Qwen3.5 the suffix \texttt{" /no\_think"} is appended to suppress the reasoning block on the tool-calling path.

\begin{spverbatim}
You are a problem-solving agent with access to tools. You operate under a strict token budget, so act efficiently. Call AT MOST ONE tool per turn -- wait to see the result before deciding what to do next. Each turn, either call one tool or call submit_answer. Think briefly, then act.
\end{spverbatim}

The task is framed as follows, with \texttt{<B>}, \texttt{<T>}, and the problem text substituted in. Note that the prompt states only that the tools exist and may be called, without revealing what any of them do.

\begin{spverbatim}
Solve the following problem within a STRICT BUDGET of <B> tokens (every token you read and generate counts toward it). PROBLEM: <problem text> You have <T> tools available (plus submit_answer). You may call any of them, and you may also work the problem out yourself. When you have the answer, call submit_answer with it. You get ONE submission: correct = win, wrong = lose.
\end{spverbatim}

The dividend and divisor are drawn uniformly at random at the digit counts fixed by the difficulty $d$ defined in Section~\ref{subsec:mcp-details}, so each additional digit adds a subtract-and-bring-down step to the manual solve. The problem statement is \texttt{"Compute the integer quotient and remainder of the long division <dividend> / <divisor>. Give the quotient and the remainder."}, and the checker accepts any answer whose first two integers match the quotient and remainder (a bare quotient suffices when the remainder is zero).

The meter is a single monotonic token count accrued after every model call, summing output tokens and billable input tokens, and for Anthropic cache-read and cache-write tokens are folded back into input so the meter is cache-invariant and comparable across providers. The live budget is only a safety stop, and the true win/lose threshold is applied post hoc, so the budget and the flat per-call surcharge (default $25$ tokens) are free re-scoring axes. The per-cell budget is $\mathrm{budget}(d, T) = \big(900 + 90\,\max(0, d-8)\big) + 1800\,T$.

Each family was run over $T \in \{4, 8, 12, 16, 20\}$ tools and difficulty $d \in \{8, 10, 12, 14, 16\}$, with repetitions per cell scaling inversely with $T$ (\{$4{:}15$, $8{:}9$, $12{:}6$, $16{:}3$, $20{:}3$\}) so that each $(T, d)$ cell yields a comparable number of scored oracle-assignment rows, and rep $r$ uses seed $r$ shared across cells so a given rep is the same problem across models and tool counts.

\section{Derivation of \textit{Lomekwi} Action Budget}
\label{app:cost}

Here we derive the budget formulas used in Section \ref{subsec:lom-details}. In \textit{Lomekwi}, grinding the \textit{Examine}-\textit{Use} path to probabilistically get all the keys is simply an instance of the well-known coupon collector problem \citep{flajolet1992coupon}. It is well known that for an instance of this problem with $n$ coupons (or, in this case, keys), the expected number of turns needed to obtain all the items is $nH_n\sim n\log n$. Since \textit{Lomekwi} further requires that each key is \textit{Used} on the corresponding door, the total expected time for the grind path is $\mathbb{E}[\text{grind}]=nH_n+n.$

\begin{figure}
  \centering
  \includegraphics[width = 0.5\linewidth]{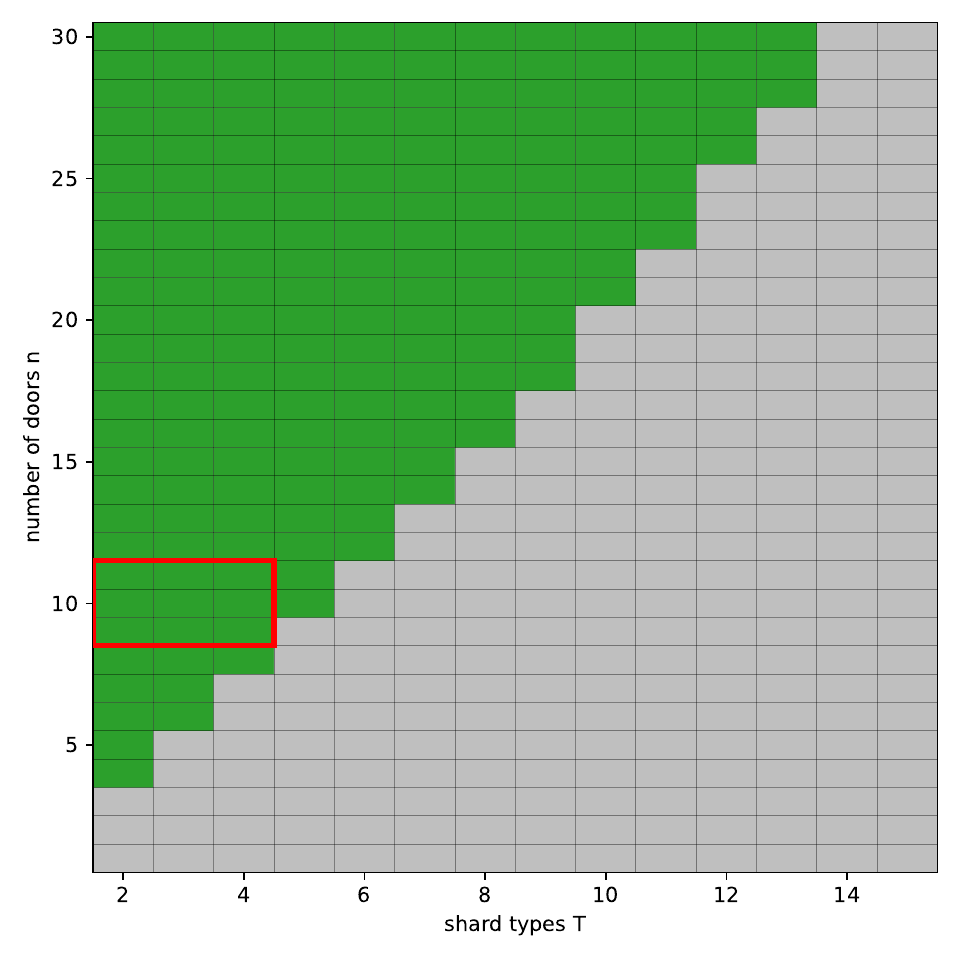}
  \caption{A large region parameterized by $T$ and $n$. Green means $\mathbb{E}[\text{build}]<\mathbb{E}[\text{grind}]$. We conduct our tests in the red box.}
  \label{fig:app-budget-vs-n}
\end{figure}

The build path is slightly more complex to analyze. Because the agent accumulates keys concurrently with shards, the exact optimal policy is unnecessarily complicated. Instead, we assume that the build path requires the agent to complete three steps sequentially: collecting the shards needed to create the machine, trying combinations of shards until the machine is built, and then using the machine to obtain the remaining keys. The expected length of the build path under this assumption serves as a reasonable upper bound for the expectation under optimal play. Since we only simulate instances of \textit{Lomekwi} where the build path remains optimal under this assumption (see Figure \ref{fig:app-budget-vs-n}), this does not affect our results.

Algebraically, we write $\mathbb{E}[\text{build}]=\mathbb{E}[\text{collect}]+\mathbb{E}[\text{combine}]+\mathbb{E}[\text{open}]$. The first term is another instance of the coupon collector problem, this time with $T$ coupons (shards). After all $T$ shards are collected, finding the correct pair takes an expected $\frac{{T\choose 2}+1}{2}$ \textit{Combine} actions. Once the machine is obtained, the agent must then \textit{Use} it to obtain all the keys it didn't get while \textit{Examining} for the shards. Again, we trivially bound this expectation by $2n$. This yields $$\mathbb{E}[\text{collect}]=TH_T,\quad \mathbb{E}[\text{combine}]=\frac{{T\choose 2}+1}{2},\quad \mathbb{E}[\text{open}]\leq 2n.$$

Figure \ref{fig:app-budget-vs-n} shows a large region parameterized by $T$ and $n$. If $\mathbb{E}[\text{build}]\geq \mathbb{E}[\text{grind}]$ in a region, it is colored gray. Otherwise, it is colored green. Our tests occur in the region defined by $T\in\{2,3,4\}$,\,$n\in\{9,10,11\}$, outlined in red. Building is theoretically optimal to grinding in this region, so building the tool presents some objective benefit. 
      
As stated in Section \ref{subsec:lom-details}, we also impose an \textit{action budget} of $B=\lfloor1.2\mathbb{E}[\text{grind}]\rfloor$. The existence of an action budget is justified in Section \ref{subsec:exp-setup}, and we choose to scale $B$ with $\mathbb{E}[\text{grind}]$ so that grinding is always feasible. The constant factor $1.2$ is lightly tuned; from \citep{flajolet1992coupon}, the probability of collecting all $n$ keys with $b$ \textit{Examine} actions is $$
\Pr(\text{all } n \text{ keys collected within } b \text{ draws}) = \sum_{k=0}^{n} (-1)^k \binom{n}{k} \left(\frac{n-k}{n}\right)^{b}. $$Within the region we consider (Figure \ref{fig:app-budget-vs-n}), setting $b=B-n$ (to reserve $n$ \textit{Use} actions) and evaluating the formula above yields a high but risky probability of success. For example, with $n=10$, we have $B=47$ and $$\Pr(\text{all } 10 \text{ keys collected within } 37 \text{ draws})\approx 80.9\%.$$This appears feasible but is inherently risky. On the other hand, since $T$ is very low in the region we consider ($T\leq 4$ throughout), it is much more probable to succeed under building. For example, consider the square $(T,n)=(4,10)$. Even assuming it takes the maximum of ${4\choose 2}=6$ \textit{Combine} actions to make the machine, the agent would have at least $B-{T\choose 2}-2n=47-6-20=21$ turns to find $4$ different shards. Substituting $n=4$ and $b=21$ into the success formula yields $$\Pr(\text{all } 4 \text{ shards collected within } 21 \text{ draws})\approx 99.1\%.$$Though grinding is possible within the region, building is far more reliable.

\section{Obfuscation Choices}
\label{app:obfuscation}

As motivated in Section~\ref{subsec:prior-suppression}, we relabel every latent role per episode to suppress pretraining priors. The relabeling protocol is fixed (a fresh uniform-random assignment averaged over episodes), but the \emph{pool} the labels are drawn from is a design choice, and it turns out to matter a great deal. We implemented three schemes, summarized with example labels in Table~\ref{tab:obf-schemes} (Appendix~\ref{app:env}), and tested them on \emph{Lomekwi} in a matched sweep (Claude Opus, 40 episodes per scheme, identical worlds).

The first scheme draws pronounceable nonsense words such as \texttt{mifkand}, \texttt{skond}, and \texttt{varmkey}. Because these read like invented proper nouns, they are the most natural-looking obfuscation, but they proved actively harmful. The word-like tokens invited the model to confabulate an entire fantasy game around them. In one representative Opus episode the model fabricated a hint that the environment never emitted, ``\emph{THREE AS ONE, fuse the trio into a masterglyph},'' and then proceeded to invent a ``rollback ward,'' a ``rune-sphere'' that ``resets the vault,'' and a ``wardbane'' to destroy it, none of which exist in \emph{Lomekwi}. The evocative labels supplied enough narrative scaffolding for the model to hallucinate mechanics and even fake its own victory messages, so although episodes sometimes still solved by brute force ($24/40$), the traces were dominated by invented structure rather than genuine discovery.

The second scheme draws random \texttt{[a-zA-Z0-9]} strings such as \texttt{5GOSZE6r}. These carry no connotation at all, which fixes the confabulation problem, but they overshoot. The labels are so hard to track and reproduce that the model loses the thread of what it holds, misremembers items, and issues actions on objects that do not exist. This scheme solved $0/40$ episodes with elevated no-op rates, failing not because the task was hard but because the model could not reliably manipulate the tokens.

The third scheme labels each role with a single distinct letter (\texttt{k}, \texttt{x}, \texttt{c}, and so on). It inherits the connotation-freeness of the alphanumeric scheme, since a bare letter primes no crafting narrative, while remaining trivially easy to track and copy back exactly, which avoids the manipulation failures. On the matched sweep it solved $26/40$ episodes, the highest of the three, with no confabulation and no tracking errors. We therefore use single-letter obfuscation for all reported \emph{Lomekwi} and recognition-MCP runs. The lesson is that obfuscation faces a two-sided constraint: labels must be semantically empty enough not to prime a prior (ruling out pronounceable words), yet simple enough to manipulate reliably (ruling out random strings), and single letters are the sweet spot.

\section{Notable Variations of \textit{Lomekwi}}
\label{app:variations}

\emph{Deep Lomekwi} adds variation to the base \emph{Lomekwi} game by introducing depth of tool creation. In the same setting as \emph{Lomekwi}, instead of combining two shards to create the machine, they now combine to create a "supershard". Two such supershards then combine to make the machine, meaning that if some shard $x$ and some shard $y$ combine to create the supershard, then building the machine requires 2 $x$'s and 2 $y$'s.

\begin{figure}
    \centering
    \includegraphics[width=0.7\linewidth]{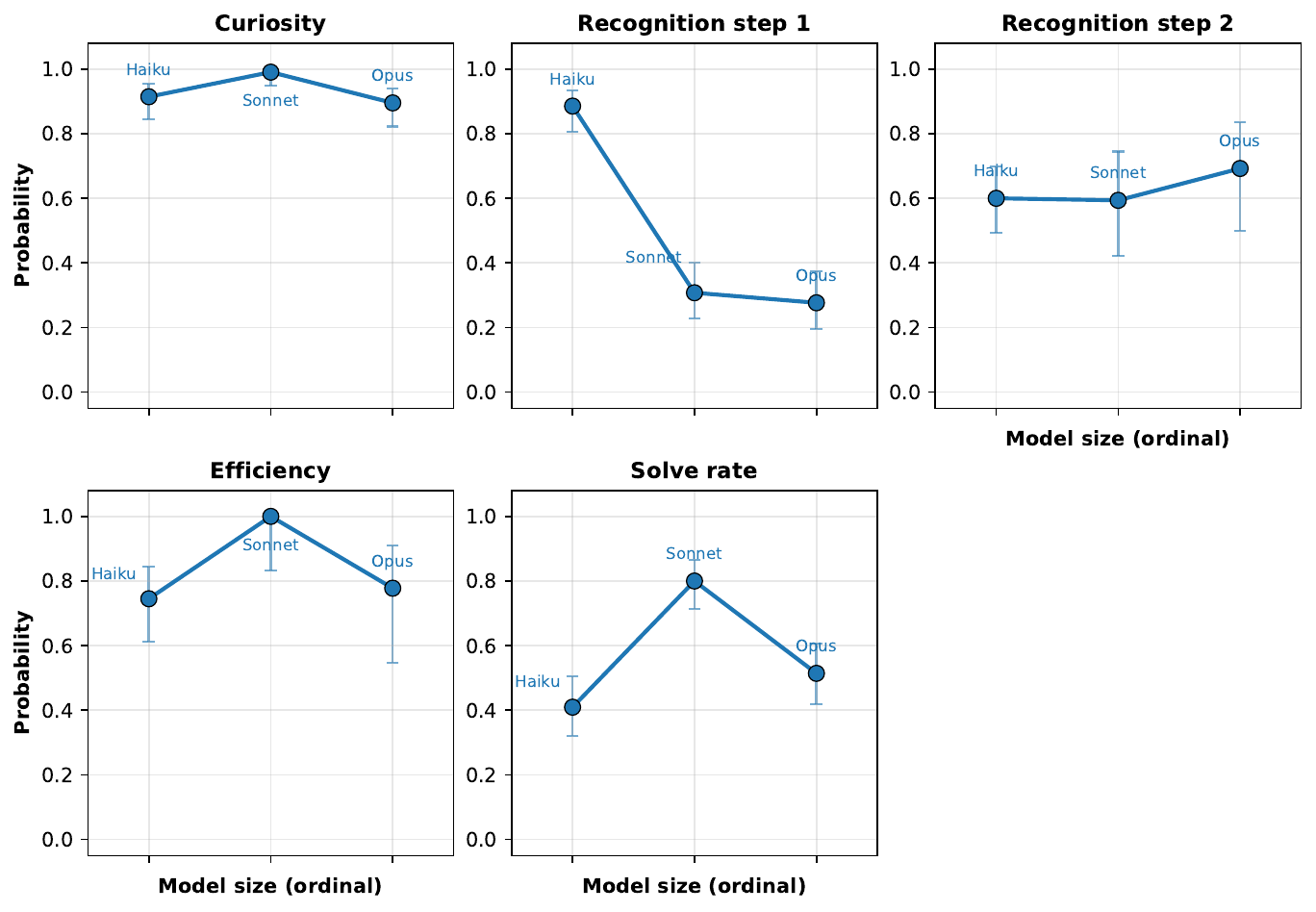}
    \caption{Each panel is one step, vs.\ model size (ordinal)}
    \label{fig:deep-lomekwi}
\end{figure}

By introducing depth, recognition $R$ can now be split into two parts that represent the two steps of tool creation: $R_1$ ($\Pr(\text{Build supershard} \mid \text{Both shards})$) and $R_2$ ($\Pr(\text{Build machine} \mid \text{Supershard})$)  which multiply to make $R$. As seen in Figure~\ref{fig:deep-lomekwi}, the inversion of recognition present in standard \emph{Lomekwi} is shown in $R_1$ where the larger models (Claude Sonnet and Opus) perform drastically worse than the smaller model (Claude Haiku). Additionally, $R_2$ is relatively flat as models grow larger, resulting in the overall recognition $R$ inverting with model size. We conjecture that the flatness of $R_2$ is due to selection bias: when Sonnet and Opus recognize to build the supershard, these model instances are skewed to be more willing to recognize future builds as well.

\end{document}